\documentclass[runningheads]{llncs}
\usepackage{graphicx}

\usepackage{tikz}
\usepackage{comment}
\usepackage{amsmath,amssymb} 
\usepackage{color}

\usepackage{pifont}%
\begin{document}
\pagestyle{headings}
\mainmatter
\def\ECCVSubNumber{100}  

\title{A Study on Self-Supervised Object Detection Pretraining} 

\titlerunning{ECCV-22 submission ID \ECCVSubNumber} 
\authorrunning{ECCV-22 submission ID \ECCVSubNumber} 
\author{Anonymous ECCV submission}
\institute{Paper ID \ECCVSubNumber}

\titlerunning{A Study on Self-Supervised Object Detection Pretraining}
%
\author{Trung Dang\inst{1} \and
Simon Kornblith\inst{2} \and
Huy Thong Nguyen\inst{2} \and
Peter Chin\inst{3}
Maryam Khademi\inst{2}}
\authorrunning{Dang et al.}
%
\institute{Boston University \and
Google LLC \and Dartmouth College}
\maketitle

\begin{abstract}
In this work, we study different approaches to self-supervised pretraining of object detection models. We first design a general framework to learn a spatially consistent dense representation from an image, by randomly sampling and projecting boxes to each augmented view and maximizing the similarity between corresponding box features. We study existing design choices in the literature, such as box generation, feature extraction strategies, and using multiple views inspired by its success on instance-level image representation learning techniques \cite{swav,dino}. Our results suggest that the method is robust to different choices of hyperparameters, and using multiple views is not as effective as shown for instance-level image representation learning.
We also design two auxiliary tasks to predict boxes in one view from their features in the other view, by (1) predicting boxes from the sampled set by using a contrastive loss, and (2) predicting box coordinates using a transformer, which potentially benefits downstream object detection tasks. We found that these tasks do not lead to better object detection performance when finetuning the pretrained model on labeled data.

\end{abstract}
\section{Introduction}

Pretraining a model on a large amount of labeled images and finetuning on a downstream task, such as object detection or instance segmentation, has long been known to improve both performance and convergence speed on the downstream task. Recently, self-supervised pretraining has gained popularity since it significantly reduces the cost of annotating large-scale datasets, while providing superior performance compared to supervised pretraining. Although a number of prior works obtain semantic representations from unlabeled images via the use of proxy tasks \cite{noroozi2016unsupervised,larsson2016learning,doersch2017multi}, recent works focus on instance-level contrastive learning \cite{simclr,simclr2,moco,moco-v2} or self-distillation \cite{byol,dino} methods. These methods learn an instance-level representation for each image that performs competitively with  using only a linear \cite{simclr} or K-nearest neighbor \cite{dino} classifier, closing the gap to the performance of supervised baselines.

However, most of these contrastive and self-distillation methods focus on learning an instance-level representation. They are likely to entangle information about different image pixels, and are thus sub-optimal for transfer learning to dense prediction tasks. A recent line of work aims to improve self-supervised pretraining for object detection by incorporating modules in the detection pipeline \cite{soco}, and taking into account spatial consistency \cite{vader,scrl,soco,densecl,pixpro,instaloc}. Rather than comparing representation at instance-level, these methods propose to leverage view correspondence, by comparing feature vectors \cite{densecl,vader} or RoIAlign features of sampled boxes \cite{scrl,soco} at the same location from two augmented view of the same image. These pretraining strategies have been shown to benefit downstream dense prediction tasks. However, there may still be a discrepancy between the pretraining and the downstream task, since they are optimized towards different objectives.

Following prior works that pretrain SSL models for object detection models by sampling object bounding boxes and leveraging view correspondence~\cite{scrl,soco}, we study how different design choices affect performance. Specifically, we investigate strategies for box sampling, extracting box features, the effect of multiple views (inspired by \verb!multi-crop! \cite{swav}), and the effect of box localization auxiliary tasks. We evaluate these proposals by pretraining on ImageNet dataset and finetuning on COCO dataset. Our results suggest that (1) the approach is robust to different hyperparameters and design choices, (2) the application of \texttt{multi-crop} and box localization pretext tasks in our framework, as inspired by their success in the literature, does not lead to better object detection performance when finetuning the pretrained model on labeled data. 


\section{Related Work}

\subsection{Self-Supervised Learning from Images}


A large number of recent work on self-supervised learning focuses on constrastive learning, which learns the general feature of an image by using data augmentations and train the model to discriminate views coming from the same image (positive pairs) and other images (negative pairs), usually by using the InfoNCE loss \cite{gutmann2010noise,wu2018unsupervised,kihyunk2016advances}. In practice, contrastive learning requires simultaneous comparison among a large number of sampled images, and benefits from large batches \cite{simclr,simclr2}, memory banks \cite{moco,moco-v2}, false negative cancellation \cite{Huynh_2022_WACV}, or clustering \cite{caron2018deep,asano2019self,huang2019unsupervised}. Recently, non-contrastive methods such as BYOL \cite{byol} and its variants \cite{dino,siamese} obtain competitive results without using negative pairs, by training a student network to predict the representations obtained from a teacher network.

While instance-level representation learning methods show promising results on transfer learning for image classification benchmarks, they are sub-optimal for dense prediction tasks. A recent line of works focus on pre-training a backbone for object detection. Similar to instance-level representation learning, these pre-training methods can be based on contrastive learning (e.g. VADeR \cite{vader}, DenseCL \cite{densecl}, and PixPro \cite{pixpro}),
or self-distillation, (e.g. SoCo \cite{soco} and SCRL \cite{scrl}). These methods share a general idea of leveraging view correspondence, which is available from the spatial relation between augmented views and the original image. Beyond pre-training the backbone, UP-DETR \cite{up-detr} and DETReg \cite{detreg} propose a way to pre-train a whole object detection pipeline, using a frozen backbone trained with SSL on object-centric images to extract object features. Xie et al. \cite{orl} leverage image-level self-supervised pre-training to discover object correspondence and perform object-level representation learning from scene images.

Many of these object detection pre-training techniques rely on heuristic box proposals \cite{soco,detreg} or frozen backbone trained on ImageNet \cite{up-detr,detreg}. Our work, however, aims to study the potential of end-to-end object detection pre-training without them. The framework we study is closest to SCRL \cite{scrl}, which adopts the approach from BYOL \cite{byol}.

\subsection{Object Detection}

\paragraph*{Faster R-CNN} \cite{fasterrcnn} is a popular object detector, which operates in two stages. In the first stage, a single or multi-scale features extracted by the backbone are fed into the Region Proposal Network to get object proposals. In the second stage, the pooled feature maps inside each bounding box are used to predict objects. Low-quality object proposals and predictions are filtered out with Non-Maximum Suppression (NMS), which is heuristic and non-differentiable.

\paragraph*{Detection Transformer (DETR)} \cite{detr} is a simpler architecture than Faster R-CNN and operates in a single stage. The features retrieved by the backbone are encoded by a transformer encoder. The transformer decoder attends to the encoded features and uses a fixed number of query embeddings to output a set of box locations and object categories. DETR can learn to remove redundant detections without relying on NMS; however, the set-based loss and the transformer architecture are notoriously difficult to train. Deformable DETR \cite{ddetr} with multi-scale deformable attention modules has been shown to improve over DETR in both performance and training time.
\section{Approach}
\label{sec:notations}

In this section, we describe the general framework and the notations we use in our study. We first generate multiple views of an image via a sequence of image augmentations. Our framework aims to learn a spatially consistent representation by matching features of boxes covering the same region across views. To avoid mode collapse, we train a student network to predict the output of a teacher network, following BYOL \cite{byol}. At the end of this section, we compare the proposed framework with a number of existing pretraining techniques for object detection.

\subsection{View Construction and Box Sampling}We first randomly crop the original image to obtain a base view, with minimum scale of $s_{\text{base}}$. Next, $V$ augmented views $v_1,\cdots v_V$ are constructed from the base view using $V$ different image augmentations $t_{1},\cdots, t_{V}\sim\mathcal{T}$, each of which is a sequence of random cropping with minimum scale of $s_{\text{view}}$, color jittering, and Gaussian blurring. Here, $\mathcal{T}$ is the distribution of image augmentations. The minimum scale of these $V$ views with regards to the original image is $s_{\text{base}}\times s_{\text{view}}$. We separate $s_{\text{base}}$ and $s_{\text{view}}$ and choose $s_{\text{view}}>0.5$ to make sure that views are pairwise overlapped. The views are also resized to a fixed size to be processed in batches.

Next, we sample $K$ boxes $b^1,\cdots,b^{K}\sim\mathcal{B}$ relative to the base view, where $b^k\in\mathbb{R}^4$ is the box coordinate of the top left and the bottom right corner, $\mathcal{B}$ is the box distribution. We transform these boxes from the base view to each augmented view $v_i$, based on the transformation $t_i$ used to obtain $v_i$. We keep only valid boxes that are completely fitted inside the view. Let $\tau_{t_i}$ be the box transformation, $B_i$ be the set of valid box indices by this transformation, the set of boxes sampled for view $i$ is denoted as $\{b^k_i=\tau_{t_i}(b^k)|k\in B_i\}$. During this sampling process, we make sure that each box in the base view is completely inside at least two augmented views by over-sampling and removing boxes that do not satisfy this requirement.

\begin{figure}[t]
    \centering
    \includegraphics[width=0.8\textwidth]{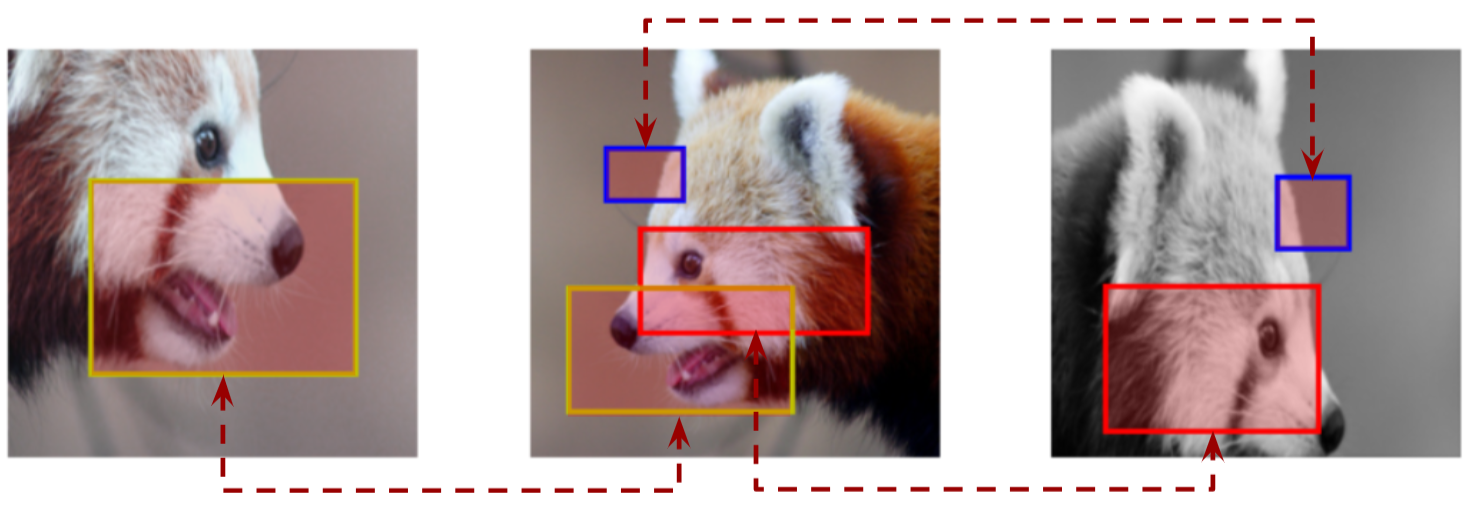}
    \caption{Example of view and random box generation with number of views $V=3$ and number of boxes $K=3$.}
    \label{fig:aug}
\end{figure}

The result of this view construction and box sampling process is $V$ augmented views of the original image with a set of at most $K$ boxes for each view. Figure \ref{fig:aug} describes the view and random box generation with $V=3$ and $K=3$.

\subsection{SSL Backbone}Our strategy for training the backbone follows BYOL~\cite{byol}. We train an online network $f_{\theta}$, whose output can be used to predict the output of a target network $f_{\xi}$, where $\theta$ and $\xi$ are weights of the online and target network, respectively. The target network is built by taking an exponential moving average of the weights of the online network, which is analogous to model ensembling and has been previously shown to improve performance~\cite{byol,dino}. 

Specifically, let $y_i=f_{\theta}(v_i), y'_i=f_{\xi}(v_i)$ be the output of the backbone for each view $i$. Different from instance-level representation learning \cite{byol,dino} where global representations are compared, we compare local regions of interest refined by sampled boxes. Let $\phi$ be a function that takes a feature map $y$, and box coordinates $b$ to output a box representation $\phi(y,b)$ (e.g., RoIAlign). Following \cite{byol}, we add projection layers $g_\theta, g_\xi$ for both networks and a prediction layer $q_\theta$ for the online network. We obtain box representations $u_i^k=g_\theta(\phi(y_i,b_i^k))$ from the online network, and ${u'}_i^k=g_\xi(\phi(y'_i,b_i^k))$ from the target network.
The SSL loss is computed as:

\begin{equation}\mathcal{L}_{\text{BYOL}}(y_i, y'_i, b_i; \theta)=\sum_{\substack{i=1,j=1\\j\ne i}}^V\frac{1}{|B_i\cap B_j|}\sum_{k\in B_i\cap B_j}\|q_\theta(u_i^k)-{u'}_j^k\|^2\label{eqn:byol}\end{equation}

\subsection{Comparison with Prior Work}

Our framework reduces to BYOL when $V=2$, $K=1$, and $\tau_{t_i}$ returns the whole view boundary regardless of the transformation $t_i$. In this case, only global representation of two augmented views are compared.

The framework is similar to SCRL when $V=2$, $\mathcal{B}$ is a uniformly random box distribution, and $\phi(y, b)$ outputs the $1\times 1$ RoIAlign of the feature map $y$ with regards to box $b$. We do not remove overlapping boxes as in SCRL, since object bounding boxes are not necessarily separated.

In SoCo \cite{soco}, boxes are sampled from box proposals via the selective search algorithm. A third view (and fourth view), which is a resizing of one of the two views is also included to encourage learning object scale variance. We generalize it to $V$ views in our framework. SoCo offers pretraining FPN layers in Faster R-CNN for better transfer learning efficiency. We however only focus on pretraining the ResNet backbone, which is included in both Faster R-CNN and DETR.


\begin{figure}[t]
    \centering
    \includegraphics[width=\textwidth]{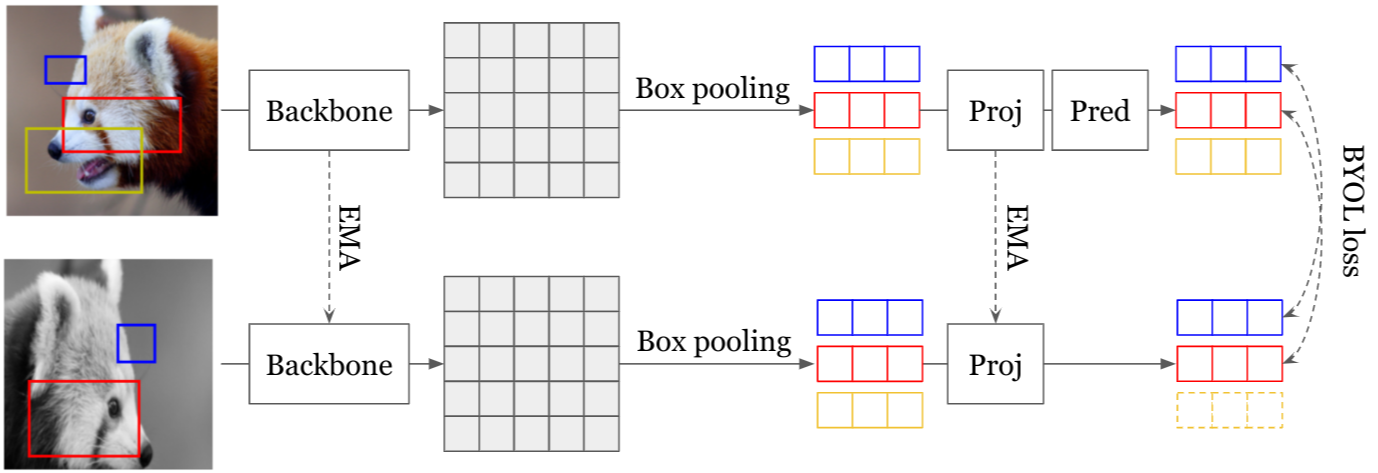}
    \caption{General framework: we train an online network to predict the output of a target network under a different augmented view and update the target network with a moving average of the online network. The representations are compared at regional level via random box sampling.}
    \label{fig:common}
\end{figure}
\section{Experiments}

\subsection{Experimental Setup}
\label{sec:setup}

\paragraph*{Dataset}We pretrain the ResNet50 backbone on ImageNet \cite{deng2009imagenet} ($\sim$ 1.28m images), and finetune the object detection model on MS-COCO \cite{lin2014microsoft} ($\sim$ 118k images).

\paragraph*{Image Augmentation}After we crop a view from the image as described in Section \ref{sec:notations}, we resize it to $256\times 256$, and follow previous work \cite{simclr,byol,scrl} in applying random horizontal flipping, color jittering, and Gaussian blurring.

\paragraph*{Network Architecture}We use ResNet-50 \cite{resnet} as the backbone, which outputs a feature map of shape $(7, 7, 2048)$. 

\paragraph*{Pretraining Setup} For the baseline, we follow \cite{byol} to train a BYOL model for 300 epochs (top-1: 73.0) and 1,000 epochs (top-1: 74.3). For our framework, if not explicitly stated otherwise, we use $V=2, K=8, s_{\text{base}}=0.9, s_{\text{view}}=0.6$. We use \texttt{LARS} optimizer with a base initial learning rate $0.3\times\text{batch size}/256$ for 300 epochs and $0.2\times\text{batch size}/256$ for 1,000 epochs, with a cosine learning rate decay, and a warm up period of 10 epochs.

\paragraph*{Evaluation} We evaluate pretrained models on the COCO object detection task. We fine-tune a Mask R-CNN detector with an FPN backbone on the COCO train2017 split with the standard $1\times$ schedule, following \cite{he2019rethinking}. For DETR fine-tuning, we use the initial learning rate $1\times 10^{-4}$ for transformers and $5\times 10^{-5}$ for the CNN backbone, following UP-DETR \cite{up-detr}. The model is trained with 150 and 300 epoch schedules, with the learning rate multiplied by 0.1 at 100 and 200 epochs, respectively.

Table \ref{table:baseline} compares our framework with both instance-level and dense pretraining methods. Our proposed framework shows a clear performance improvement over methods that only consider instance-level contrasting. Among methods that leverage view correspondence to learn dense representation, our results are comparable with SCRL. Note that some methods, for example DetCon and SoCo, use unsupervised heuristic in obtaining object bounding boxes or segmentation masks, thus are not directly comparable.

\begin{table}
\begin{center}

\caption{Results of object detection and instance segmentation fine-tuned on COCO with Faster R-CNN + FPN.}
\label{table:baseline}
\begin{tabular}{ll|c|ccc}
\hline\noalign{\smallskip}
& Method & Epoch & AP$^b$ & AP$_{50}^b$ & AP$_{75}^b$\\
\noalign{\smallskip}
\hline
\noalign{\smallskip}
& random init.  & - & 32.8 & 50.9 & 35.3 \\ 
& supervised & - & 39.7 & 59.5 & 43.3 \\ 
\hline
& MoCo & 200 & 38.5 & 58.3 & 41.6 \\ 
& SimCLR \cite{simclr} & - & 38.5 & 58.0 & 42.0 \\ 
& BYOL$^*$ \cite{byol} & 300 & 39.1 & 59.9 & 42.4 \\
& BYOL$^*$ \cite{byol} & 1k & 40.1 & 61.3 & 43.9 \\ 
& MoCo-v2 & - & 39.8 & 59.8 & 43.6 \\ 
\hline
& DetCon$_S$ \cite{detcon} & 300 & 41.8 & - & - \\ 
& DetCon$_B$ \cite{detcon} & 300 & 42.0 & - & - \\ 
& SCRL \cite{scrl} & 1k & 40.9 & 62.5 & 44.5 \\
& SoCo \cite{soco} & 100 & 42.3 & 62.5 & 46.5 \\
& SoCo \cite{soco} & 400 & 43.0 & 63.3 & 47.4 \\
& DenseCL \cite{densecl} & 200 & 40.3 & 59.9 & 44.3 \\ 
 & PLRC \cite{plrc} & 200 & 40.7 & 60.4 & 44.7 \\
\hline
& Ours & 300 & 39.9 & 60.7 & 43.8 \\
& Ours & 1k & 40.8 & 62.1 & 44.7 \\
\hline
\end{tabular}
\end{center}
\end{table}

In the following sections, we explore different settings and techniques built on top of this framework to study if they improve the performance.

\subsection{Effect of Box Sampling Strategies}

We focus on a general random box sampling strategies, as in \cite{scrl,up-detr}. While some box proposal algorithms (e.g., selective search) have been shown to produce sufficiently good object boundaries to improve SSL performance \cite{soco,up-detr}, we want to avoid incorporating additional inductive bias in the form of rules to generate boxes, since the efficacy of such rules could depend on the dataset.

We study the effect of four hyperparameters of the random box sampling strategy: (1) number of boxes per image ($K$), (2) box coordinate jittering rate (relative to each coordinate value) ($\%n$), (3) minimum box size ($S_{\text{min}}$), and (4) minimum scale for each view ($s_{\text{view}})$. For each attribute, we report results on several chosen values as in Table \ref{table:box_sampling}.

The results are shown in \ref{table:box_sampling}. For the number of boxes per image $K$, it can be seen that increasing the number of boxes does not have a large effect on the finetuning performance. The results slightly drop when introducing box jittering, which was proposed in \cite{soco}. The approach is pretty robust against changing the minimum box size $S_{\text{min}}$. For the minimum scale for each view $s_{\text{view}}$, it can be observed that having a larger scale (i.e. larger overlapping area between views, which boxes are sampled from) does not help to increase the pretraining efficacy.

\begin{table}[h]
\begin{center}
\caption{Effect of Box Sampling by number of boxes $K$, box jittering rate $\%n$, minimum box size $S_{\min}$, and minimum scale for each view $s_{\text{view}}$. Underlined numbers are results of the default setting (Section \ref{sec:setup}).}
\vspace{-5mm}
\label{table:box_sampling}
\begin{tabular}{cccc|cccc|cccc|cccc}
\hline\noalign{\smallskip}
 $K$ & AP$^b$ & AP$_{50}^b$ & AP$_{75}^b$ & $\%n$ & AP$^b$ & AP$_{50}^b$ & AP$_{75}^b$ & $S_{\text{min}}$ & AP$^b$ & AP$_{50}^b$ & AP$_{75}^b$ & $s_{\text{view}}$ & AP$^b$ & AP$_{50}^b$ & AP$_{75}^b$\\
\noalign{\smallskip}
\hline
\noalign{\smallskip}
  4 & 40.0 & 60.8 & 43.7 & 0 & \textbf{\underline{39.9}} & \underline{60.7} & \underline{43.8} & 0 & \underline{39.9} & \underline{60.7} & \underline{43.8} & 0.5 & \textbf{40.1} & 60.9 & 43.5 \\ 
  8 & \underline{39.9} & \underline{60.7} & \underline{43.8} & 0.05 & 39.7 & 60.5 & 43.5 & 0.05 & \textbf{40.0} & 60.7 & 43.8 & 0.6 & \underline{39.9} & \underline{60.7} & \underline{43.8} \\ 
  16 & \textbf{40.2} & 61.0 & 44.1 & 0.10 & 39.7 & 60.6 & 43.1 & 0.10 & 39.9 & 60.6 & 43.3 & 0.7 & 39.6 & 60.1 & 43.2
  \\ 
  32 & 39.9 & 60.7 & 43.5 & 0.20 & 39.8 & 60.7 & 43.4 & 0.20 & 39.8 & 60.4 & 43.5 & 0.8 & 39.7 & 60.2 & 43.6 \\ 
\hline
\end{tabular}
\end{center}
\end{table}

\subsection{Effect of Methods to Extract Box Features}

We explore three different ways to extract features for each box (choices of $\phi(y, b)$): (1) RoIAlign $1\times 1$ (denoted as \texttt{ra1}), (2) RoIAlign $c\times c$ with crop size $c > 1$ (denoted as \texttt{ra3}, \texttt{ra7}, etc.), and (3) averaging cells in the feature map that overlap with the box, similar to $1\times 1$ RoIPooling (denoted as \texttt{avg}). While SCRL and SOCO use \texttt{ra1}  \cite{scrl,soco}, \texttt{ra7} offers more precise features and is used in Faster R-CNN \cite{fasterrcnn} to extract object features. \texttt{avg} shifts box coordinates slightly, introducing variance in the scale and location.

Additionally, with the use of RoIAlign $c\times c$, we want to examine the necessity of random box sampling. Specifically, we compare the dense features of the shared area of two views, which is similar to comparing $c\times c$ identical boxes forming a grid in the shared area.

The results are shown in Table \ref{table:box_features}. We observe that $\texttt{ra1}$ achieves the best performance, although the differences are marginal. Moreover, when not using random box sampling, AP scores drop significantly, hinting that comparing random boxes with diversified locations and scales is necessary for a good pretrained model.

\begin{table}[h]
\vspace{-5mm}
\begin{center}
\caption{Effect of extracting box features.}
\label{table:box_features}
\begin{tabular}{c|ccc|ccc}
\hline\noalign{\smallskip}
&& box sampling &&& shared area &\\\hline
 feature extraction & AP$^b$ & AP$_{50}^b$ & AP$_{75}^b$ & AP$^b$ & AP$_{50}^b$ & AP$_{75}^b$\\
\noalign{\smallskip}
\hline
\noalign{\smallskip}
  \texttt{ra1} & \textbf{39.9} & 60.7 & 43.8 & - & - & - \\ 
  \texttt{ra3} & 39.7 & 60.4 & 43.4 & 39.3 & 59.9 & 42.8 \\ 
  \texttt{ra7} & 39.6 & 60.4 & 43.4 & 39.4 & 60.0 & 42.8 \\ 
  \texttt{avg} & 39.8 & 60.7 & 43.3 & - & - & - \\ 
\hline
\end{tabular}
\end{center}
\vspace{-10mm}
\end{table}

\subsection{Effect of Multiple Views}

\texttt{multi-crop} has been shown to be an effective strategy in instance-level representation learning \cite{swav,dino,Huynh_2022_WACV}, with both contrastive and non-contrastive methods. In the context of dense representation learning, the only similar adoption of \texttt{multi-crop} we found is in SoCo \cite{soco}; however, their third view is only a resize of one of two main views. We are interested in examining if an adaptation of \texttt{multi-crop} that more closely resembles the original proposal of \cite{swav} provides meaningful improvements for our task. We consider two settings for our experiments:

\paragraph*{Using multiple views} In the 2-view ($V=2$) setting, since only features corresponding to the shared area of two views are considered for training, the computation related to the non-overlapping area may be useless. For better efficiency, we consider using more than two views ($V>2$). The similarity between each pair of views is included in the loss as in equation $\ref{eqn:byol}$. We expect increasing the number of views improves the pretrained model since the the model is trained using more augmented views within the same number of epochs.

\begin{figure}
    \centering
    \includegraphics[width=0.7\textwidth]{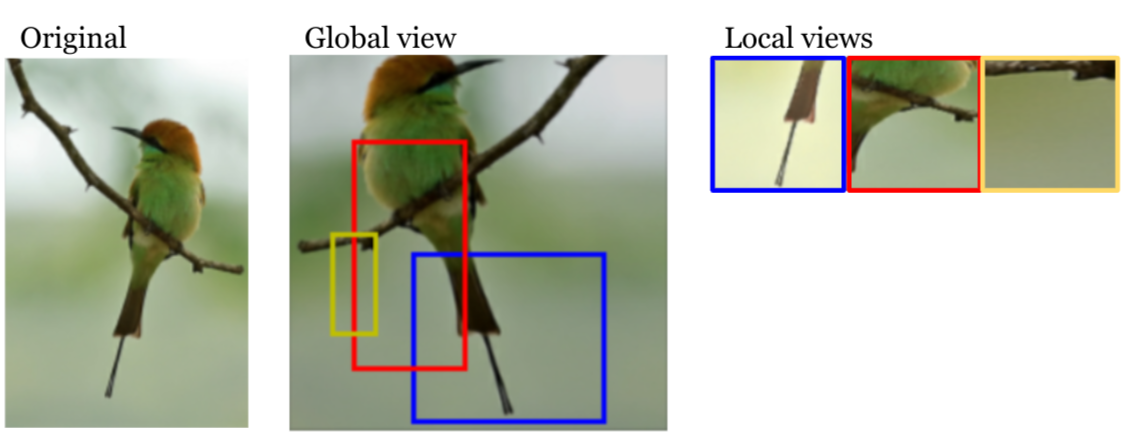}
    \caption{Leveraging local views in SSL pretraining. Each sampled box in a global view is compared with the representation of the image cropped from the box.}
    \label{fig:multicrop}
\end{figure}

\paragraph*{Using local-to-global correspondences} Instead of obtaining RoIAlign from the view's dense representation for each box, we crop the image specified by each box and obtain the features with a forward pass through the backbone. For example, if the number of boxes $K=8$, we will perform 8 forward passes through the network on the 8 crops to obtain box features. These features will be compared against box features obtained with RoIAlign from the \textit{global view}. This is similar to the adoption of local views in DINO \cite{dino}, except the local views are compared only against the corresponding regions in the feature map obtained from the global view, rather than a representation of the entire image. Figure \ref{fig:multicrop} shows how local views can be leveraged in SSL pretraining.

\begin{table}[h]
\begin{center}
\caption{Results with multiple views.}
\label{table:multiview}
\begin{tabular}{c|cccccccc}
\hline\noalign{\smallskip}
\#views & AP$^b$ & AP$_{50}^b$ & AP$_{75}^b$ \\
\noalign{\smallskip}
\hline
\noalign{\smallskip}
2 & \textbf{39.9} & 60.5 & 43.6 \\ %
3 & 39.8 & 60.8 & 43.4 \\ 
4 & 39.8 & 60.7 & 43.5\\ %
  \hline
\end{tabular}
\end{center}
\vspace{-5mm}
\end{table}

\begin{table}[h]
\begin{center}
\caption{Results with global and local views.}
\label{table:localview}
\begin{tabular}{l|ccc|ccccc}
\hline\noalign{\smallskip}
&  & ra1  &  & & avg &  \\ \hline
\#views & AP$^b$ & AP$_{50}^b$ & AP$_{75}^b$ & AP$^b$ & AP$_{50}^b$ & AP$_{75}^b$ \\
\noalign{\smallskip}
\hline
\noalign{\smallskip}
2 global & \textbf{39.9} & 60.7 & 43.8 & \textbf{39.8} & 60.7 & 43.3 \\
1 global + 8 local & 38.1 & 58.7 & 41.5 & 38.3 & 58.8 & 41.9 \\ 
2 global + 8 local & 38.3 & 58.8 & 41.7 & 39.0 & 59.7 & 42.7\\ 
\hline
\end{tabular}
\end{center}
\vspace{-5mm}
\end{table}

Table \ref{table:multiview} shows the results of using multiple views for $V=3$ or $V=4$, where we do not observe a significant performance gain despite more computation at each step. This suggests that constructing more augmented views at the same scale does not necessarily lead to an increase in the performance. Note that in this first design, we did not downscale views or use local views as in \cite{swav,dino}. For the second design, table \ref{table:localview} shows the results of using global and local views, with each local view covering an smaller area inside the global view and resized to $96\times 96$. We observe that the performance drops significantly. These results suggest that although the \texttt{multi-crop} strategy, either with or without considering local-to-global correspondences, is effective for learning global image features \cite{swav,dino}, it is not effective for learning dense features.

\subsection{Effect of Box Localization Auxiliary Task}

In addition to the SSL loss, we consider a box localization loss to match the objective of SSL pretraining with that of an object detection model. Existing methods usually improve pretraining for dense prediction tasks by leveraging spatial consistency; however, a self-supervised pretext task designed specifically for the object detection tasked has been less studied. UP-DETR \cite{up-detr} has demonstrated that using box features extracted by a well-trained vision backbone to predict box location helps DETR pretraining. In this section, we present our effort to incorporate such object detection pretext tasks into our pretraining framework. Two types of box localization loss $\mathcal{L}_{\text{box}}$ are considered. Given a box feature from a view, we can compute either (1) a box prediction loss, i.e. a contrastive loss that helps predict the corresponding box among up to $K$ boxes from another view; or (2) a box regression loss, an L1 distance and general IoU loss of box coordinates from another view predicted by using a transformer. The final loss is defined as $\mathcal{L}=\mathcal{L}_{\text{BYOL}} + \lambda\mathcal{L}_{\text{box}}$, where $\lambda$ is the weight of the box localization term.

\subsubsection{Box Prediction Loss}

Given a box feature $u_i^k$ from a view $i$, we want to predict which box from $\{u_j^1,\cdots u_j^K\}$ in view $j$ corresponds to $u_i^k$. This can be done by comparing the similarity between feature from each of these $K$ boxes with $u_i^k$. We use a contrastive loss to minimize the distance between positive box pairs, and maximize the distance between negative box pairs.

\[\mathcal{L}^{\text{box\_pred}}_{i,j}=-\sum_{k=1}^K\log\frac{\exp(\text{sim}(u_i^k, u_j^k)/\tau)}{\sum_{k'=1}^K\exp(\text{sim}(u_i^k,u_j^{k'})/\tau)}\]

\subsubsection{Box Regression Loss}

Inspired by the DETR architecture \cite{detr}, we employ a transformer, which takes $u_j^k$ as the query and looks over the representation of the $i$-th view to predict the box location $u_i^{k}$. The output of the transformer is a vector of size 4 for each box, representing the coordinate of the box center, its height and width. In addition to L1 loss, we also use the generalized IoU loss, which is invariant to box scales. The bounding box loss is defined as \[\mathcal{L}_{\text{box}}(\tilde{b}_i^k, {b}_{i}^k)=\lambda_{\text{giou}}\mathcal{L}_{\text{giou}}(\tilde{b}_i^k, {b}_i^k)+\lambda_{\text{box}}\|\tilde{b}_i^k - {b}_i^k\|_1\]where $\lambda_{\text{giou}}$ and $\lambda_{\text{box}}$ are weights in the loss, $\tilde{b}_i^k=\text{Decoder}(y_i, u_j^{k})$ is a predicted box. The box loss is defined as

\[\mathcal{L}_{\text{box}}=\sum_{\substack{i=1\\j=1,j\ne i}}^V\frac{1}{|B_i\cap B_j|}\sum_{k\in B_i\cap B_j}\left(\mathcal{L}_{\text{box}}(\tilde{b}_i^k,{b}_i^k)+\mathcal{L}_{\text{box}}(\tilde{b}^{k}_j,{b}^{k}_j)\right)\]

Table \ref{table:box_fasterrcnn} shows the results with two proposed losses when fine-tuning the Faster R-CNN model on the COCO dataset. It can be seen that these auxiliary losses, despite our expectation, have an adverse effect on the finetuning performance. We suggest that although these tasks encourage learning a representation that is useful for box prediction, the gap between these and a supervised task on labeled data is still significant that finetuning is not very effective.

\begin{table}[t]
\begin{center}
\caption{Results of object detection fine-tuned on COCO with Faster R-CNN + FPN.}
\label{table:box_fasterrcnn}
\begin{tabular}{l|c|ccccccc}
\hline\noalign{\smallskip}
Method & $\lambda$ & AP$^b$ & AP$_{50}^b$ & AP$_{75}^b$\\
\noalign{\smallskip}
\hline
\noalign{\smallskip}
random init. & - & 32.8 & 50.9 & 35.3\\
BYOL & - & 39.1 & 59.9 & 42.4 \\ \hline
No box loss & 0.00 & \textbf{39.9} & 60.5 & 43.6\\ \hline 
Box prediction & 0.01 & 39.8 & 60.3 & 43.6 \\
& 0.05 & 39.5 & 59.9 & 43.1 \\
& 0.10 & 39.5 & 60.0 & 43.4 \\
\hline
Box regression & 0.01 & 39.6 & 60.4 & 43.3 \\ 
& 0.05 & 39.1 & 59.4 & 42.9 \\
\hline
\end{tabular}
\end{center}
\end{table}

Table \ref{table:box_detr} shows the results when fine-tuning DETR, which shares the decoder architecture with the decoder used to obtain box regression loss. While our framework improves the fine-tuning performance (+0.8 $\text{AP}^b$) as shown in Table \ref{table:baseline}, it does not improve the results in the case of fine-tuning DETR (-0.4 $\text{AP}^b$).

\begin{table}[t]
\begin{center}
\caption{Results of object detection fine-tuned on COCO with DETR.}
\label{table:box_detr}
\begin{tabular}{l|cccc|ccc}
\hline\noalign{\smallskip}
Method & AP$^b$ & AP$_{50}^b$ & AP$_{75}^b$ \\
\noalign{\smallskip}
\hline
\noalign{\smallskip}
supervised & 37.7\footnote{\cite{up-detr}: 39.5} & 59.0 & 39.2 \\ \hline
BYOL \cite{byol} & 36.0\footnote{38.2 \cite{up-detr}} & 56.9 & 37.2 \\ 
Ours & 35.6 & 56.0 & 37.0 \\
Ours + regression & 35.7 & 55.8 & 37.5 \\
\hline
\end{tabular}
\end{center}
\end{table}
\section{Conclusion}

We studied a self-supervised pretraining approach for object detection based on sampling random boxes and maximizing spatial consistency. We investigated the effect of different box generation and feature extraction strategies. 
Moreover, we tried incorporating multi-crop and additional self-supervised object detection pretext tasks to the proposed framework. 
We found that the method is robust against different design choices.
\section{Acknowledgement}
We thank our colleagues from Google Brain Toronto and Brain AutoML, Ting Chen and Golnaz Ghiasi who provided insight and expertise that greatly assisted this research.

\clearpage
%
%
\bibliographystyle{splncs04}
\bibliography{egbib}

\begin{thebibliography}{10}
\providecommand{\url}[1]{\texttt{#1}}
\providecommand{\urlprefix}{URL }
\providecommand{\doi}[1]{https://doi.org/#1}

\bibitem{asano2019self}
Asano, Y.M., Rupprecht, C., Vedaldi, A.: Self-labelling via simultaneous
  clustering and representation learning. arXiv preprint arXiv:1911.05371
  (2019)

\bibitem{plrc}
Bai, Y., Chen, X., Kirillov, A., Yuille, A., Berg, A.C.: Point-level region
  contrast for object detection pre-training. arXiv preprint arXiv:2202.04639
  (2022)

\bibitem{detreg}
Bar, A., Wang, X., Kantorov, V., Reed, C.J., Herzig, R., Chechik, G., Rohrbach,
  A., Darrell, T., Globerson, A.: Detreg: Unsupervised pretraining with region
  priors for object detection. arXiv preprint arXiv:2106.04550  (2021)

\bibitem{detr}
Carion, N., Massa, F., Synnaeve, G., Usunier, N., Kirillov, A., Zagoruyko, S.:
  End-to-end object detection with transformers. In: European conference on
  computer vision. pp. 213--229. Springer (2020)

\bibitem{caron2018deep}
Caron, M., Bojanowski, P., Joulin, A., Douze, M.: Deep clustering for
  unsupervised learning of visual features. In: Proceedings of the European
  conference on computer vision (ECCV). pp. 132--149 (2018)

\bibitem{swav}
Caron, M., Misra, I., Mairal, J., Goyal, P., Bojanowski, P., Joulin, A.:
  Unsupervised learning of visual features by contrasting cluster assignments.
  Advances in Neural Information Processing Systems  \textbf{33},  9912--9924
  (2020)

\bibitem{dino}
Caron, M., Touvron, H., Misra, I., J{\'e}gou, H., Mairal, J., Bojanowski, P.,
  Joulin, A.: Emerging properties in self-supervised vision transformers. In:
  Proceedings of the IEEE/CVF International Conference on Computer Vision. pp.
  9650--9660 (2021)

\bibitem{simclr}
Chen, T., Kornblith, S., Norouzi, M., Hinton, G.: A simple framework for
  contrastive learning of visual representations. In: International conference
  on machine learning. pp. 1597--1607. PMLR (2020)

\bibitem{simclr2}
Chen, T., Kornblith, S., Swersky, K., Norouzi, M., Hinton, G.E.: Big
  self-supervised models are strong semi-supervised learners. Advances in
  neural information processing systems  \textbf{33},  22243--22255 (2020)

\bibitem{moco-v2}
Chen, X., Fan, H., Girshick, R., He, K.: Improved baselines with momentum
  contrastive learning. arXiv preprint arXiv:2003.04297  (2020)

\bibitem{siamese}
Chen, X., He, K.: Exploring simple siamese representation learning. In:
  Proceedings of the IEEE/CVF Conference on Computer Vision and Pattern
  Recognition. pp. 15750--15758 (2021)

\bibitem{up-detr}
Dai, Z., Cai, B., Lin, Y., Chen, J.: Up-detr: Unsupervised pre-training for
  object detection with transformers. In: Proceedings of the IEEE/CVF
  Conference on Computer Vision and Pattern Recognition. pp. 1601--1610 (2021)

\bibitem{deng2009imagenet}
Deng, J., Dong, W., Socher, R., Li, L.J., Li, K., Fei-Fei, L.: Imagenet: A
  large-scale hierarchical image database. In: 2009 IEEE conference on computer
  vision and pattern recognition. pp. 248--255. Ieee (2009)

\bibitem{doersch2017multi}
Doersch, C., Zisserman, A.: Multi-task self-supervised visual learning. In:
  Proceedings of the IEEE International Conference on Computer Vision. pp.
  2051--2060 (2017)

\bibitem{byol}
Grill, J.B., Strub, F., Altch{\'e}, F., Tallec, C., Richemond, P., Buchatskaya,
  E., Doersch, C., Avila~Pires, B., Guo, Z., Gheshlaghi~Azar, M., et~al.:
  Bootstrap your own latent-a new approach to self-supervised learning.
  Advances in Neural Information Processing Systems  \textbf{33},  21271--21284
  (2020)

\bibitem{gutmann2010noise}
Gutmann, M., Hyv{\"a}rinen, A.: Noise-contrastive estimation: A new estimation
  principle for unnormalized statistical models. In: Proceedings of the
  thirteenth international conference on artificial intelligence and
  statistics. pp. 297--304. JMLR Workshop and Conference Proceedings (2010)

\bibitem{moco}
He, K., Fan, H., Wu, Y., Xie, S., Girshick, R.: Momentum contrast for
  unsupervised visual representation learning. In: Proceedings of the IEEE/CVF
  conference on computer vision and pattern recognition. pp. 9729--9738 (2020)

\bibitem{he2019rethinking}
He, K., Girshick, R., Doll{\'a}r, P.: Rethinking imagenet pre-training. In:
  Proceedings of the IEEE/CVF International Conference on Computer Vision. pp.
  4918--4927 (2019)

\bibitem{resnet}
He, K., Zhang, X., Ren, S., Sun, J.: Deep residual learning for image
  recognition. In: Proceedings of the IEEE conference on computer vision and
  pattern recognition. pp. 770--778 (2016)

\bibitem{detcon}
H{\'e}naff, O.J., Koppula, S., Alayrac, J.B., van~den Oord, A., Vinyals, O.,
  Carreira, J.: Efficient visual pretraining with contrastive detection. In:
  Proceedings of the IEEE/CVF International Conference on Computer Vision. pp.
  10086--10096 (2021)

\bibitem{huang2019unsupervised}
Huang, J., Dong, Q., Gong, S., Zhu, X.: Unsupervised deep learning by
  neighbourhood discovery. In: International Conference on Machine Learning.
  pp. 2849--2858. PMLR (2019)

\bibitem{Huynh_2022_WACV}
Huynh, T., Kornblith, S., Walter, M.R., Maire, M., Khademi, M.: Boosting
  contrastive self-supervised learning with false negative cancellation. In:
  Proceedings of the IEEE/CVF Winter Conference on Applications of Computer
  Vision (WACV). pp. 2785--2795 (January 2022)

\bibitem{larsson2016learning}
Larsson, G., Maire, M., Shakhnarovich, G.: Learning representations for
  automatic colorization. In: European conference on computer vision. pp.
  577--593. Springer (2016)

\bibitem{lin2014microsoft}
Lin, T.Y., Maire, M., Belongie, S., Hays, J., Perona, P., Ramanan, D.,
  Doll{\'a}r, P., Zitnick, C.L.: Microsoft coco: Common objects in context. In:
  European conference on computer vision. pp. 740--755. Springer (2014)

\bibitem{noroozi2016unsupervised}
Noroozi, M., Favaro, P.: Unsupervised learning of visual representations by
  solving jigsaw puzzles. In: European conference on computer vision. pp.
  69--84. Springer (2016)

\bibitem{vader}
O~Pinheiro, P.O., Almahairi, A., Benmalek, R., Golemo, F., Courville, A.C.:
  Unsupervised learning of dense visual representations. Advances in Neural
  Information Processing Systems  \textbf{33},  4489--4500 (2020)

\bibitem{fasterrcnn}
Ren, S., He, K., Girshick, R., Sun, J.: Faster r-cnn: Towards real-time object
  detection with region proposal networks. Advances in neural information
  processing systems  \textbf{28} (2015)

\bibitem{scrl}
Roh, B., Shin, W., Kim, I., Kim, S.: Spatially consistent representation
  learning. In: Proceedings of the IEEE/CVF Conference on Computer Vision and
  Pattern Recognition. pp. 1144--1153 (2021)

\bibitem{kihyunk2016advances}
Sohn, K.: Improved deep metric learning with multi-class n-pair loss objective.
  In: Lee, D., Sugiyama, M., Luxburg, U., Guyon, I., Garnett, R. (eds.)
  Advances in Neural Information Processing Systems. vol.~29. Curran
  Associates, Inc. (2016),
  \url{https://proceedings.neurips.cc/paper/2016/file/6b180037abbebea991d8b1232f8a8ca9-Paper.pdf}

\bibitem{densecl}
Wang, X., Zhang, R., Shen, C., Kong, T., Li, L.: Dense contrastive learning for
  self-supervised visual pre-training. In: Proceedings of the IEEE/CVF
  Conference on Computer Vision and Pattern Recognition. pp. 3024--3033 (2021)

\bibitem{soco}
Wei, F., Gao, Y., Wu, Z., Hu, H., Lin, S.: Aligning pretraining for detection
  via object-level contrastive learning. Advances in Neural Information
  Processing Systems  \textbf{34} (2021)

\bibitem{wu2018unsupervised}
Wu, Z., Xiong, Y., Yu, S.X., Lin, D.: Unsupervised feature learning via
  non-parametric instance discrimination. In: Proceedings of the IEEE
  conference on computer vision and pattern recognition. pp. 3733--3742 (2018)

\bibitem{orl}
Xie, J., Zhan, X., Liu, Z., Ong, Y., Loy, C.C.: Unsupervised object-level
  representation learning from scene images. Advances in Neural Information
  Processing Systems  \textbf{34} (2021)

\bibitem{pixpro}
Xie, Z., Lin, Y., Zhang, Z., Cao, Y., Lin, S., Hu, H.: Propagate yourself:
  Exploring pixel-level consistency for unsupervised visual representation
  learning. In: Proceedings of the IEEE/CVF Conference on Computer Vision and
  Pattern Recognition. pp. 16684--16693 (2021)

\bibitem{instaloc}
Yang, C., Wu, Z., Zhou, B., Lin, S.: Instance localization for self-supervised
  detection pretraining. In: Proceedings of the IEEE/CVF Conference on Computer
  Vision and Pattern Recognition. pp. 3987--3996 (2021)

\bibitem{ddetr}
Zhu, X., Su, W., Lu, L., Li, B., Wang, X., Dai, J.: Deformable detr: Deformable
  transformers for end-to-end object detection. arXiv preprint arXiv:2010.04159
   (2020)

\end{thebibliography}
\end{document}